\documentclass[letterpaper, 10 pt, journal, twoside]{IEEEtran}
\usepackage{amsmath,amsfonts}
\usepackage{array}
\usepackage{textcomp}
\usepackage{stfloats}
\usepackage{url}
\usepackage{verbatim}
\usepackage{graphicx}
\usepackage{cite}
\usepackage{times}
\usepackage{epsfig}
\usepackage{amssymb}
\usepackage{amsmath}
\usepackage{booktabs}
\usepackage{comment}
\usepackage{multicol}
\usepackage{multirow}
\usepackage{caption}
\usepackage{subcaption} 
\usepackage{tabularx}
\usepackage{xcolor}

\makeatother

\makeatletter

\usepackage[pagebackref=true,breaklinks=true,letterpaper=true,colorlinks,bookmarks=false]{hyperref}
\usepackage[capitalize]{cleveref}

\begin{document}
%
\title{HHI-Assist: A Dataset and Benchmark of Human-Human Interaction in Physical Assistance Scenario}
%
%
%


\DeclareRobustCommand*{\IEEEauthorrefmark}[1]{%
  \raisebox{0pt}[0pt][0pt]{\textsuperscript{\footnotesize\ensuremath{#1}}}}

\author{\authorblockN{Saeed Saadatnejad\authorrefmark{*,1},
Reyhaneh Hosseininejad\authorrefmark{*,1},
Jose Barreiros\authorrefmark{2}, 
Katherine M. Tsui\authorrefmark{2} and
Alexandre Alahi\authorrefmark{1}}


}

\markboth{IEEE Robotics and Automation Letters. Preprint Version. Accepted June, 2025}
{Saadatnejad \MakeLowercase{\textit{et al.}}: HHI-Assist} 

%


\maketitle

\let\thefootnote\relax\footnotetext{
Manuscript received: December 29, 2024; Revised: March 26, 2025; Accepted: April 17, 2025. This paper was recommended for publication by
Editor Angelika Peer upon evaluation of the Associate Editor and Reviewers' comments.}

\let\thefootnote\relax\footnotetext{$^1 $Saeed Saadatnejad, Reyhaneh Hosseininejad and Alexandre Alahi are with VITA laboratory, EPFL, Lausanne, Switzerland (email: \{saeed.saadatnejad, reyhaneh.hosseininejad, alexandre.alahi\}@epfl.ch).} 
\let\thefootnote\relax\footnotetext{$^2 $Jose Barreiros and Katherine M. Tsui are with Toyota Research Institute, Cambridge, MA, USA (email: \{jose.barreiros, kate.tsui\}@tri.global).}

\let\thefootnote\relax\footnotetext{\leftline{$^* $ Equal contribution. }}

\let\thefootnote\relax\footnotetext{Digital Object
Identifier (DOI): see top of this page.
}

\begin{abstract}

The increasing labor shortage and aging population underline the need for assistive robots to support human care recipients. To enable safe and responsive assistance, robots require accurate human motion prediction in physical interaction scenarios. However, this remains a challenging task due to the variability of assistive settings and the complexity of coupled dynamics in physical interactions.
In this work, we address these challenges through two key contributions: (1) \textbf{HHI-Assist}, a dataset comprising motion capture clips of human-human interactions in assistive tasks; and (2) a conditional Transformer-based denoising diffusion model for predicting the poses of interacting agents.
Our model effectively captures the coupled dynamics between caregivers and care receivers, demonstrating improvements over baselines and strong generalization to unseen scenarios. By advancing interaction-aware motion prediction and introducing a new dataset, our work has the potential to significantly enhance robotic assistance policies.
The dataset and code are available at:
\href{https://sites.google.com/view/hhi-assist/home}{https://sites.google.com/view/hhi-assist/home}.
\end{abstract}


\section{Introduction}

\begin{figure}[!t]
    \centering
    \includegraphics[width=\linewidth]{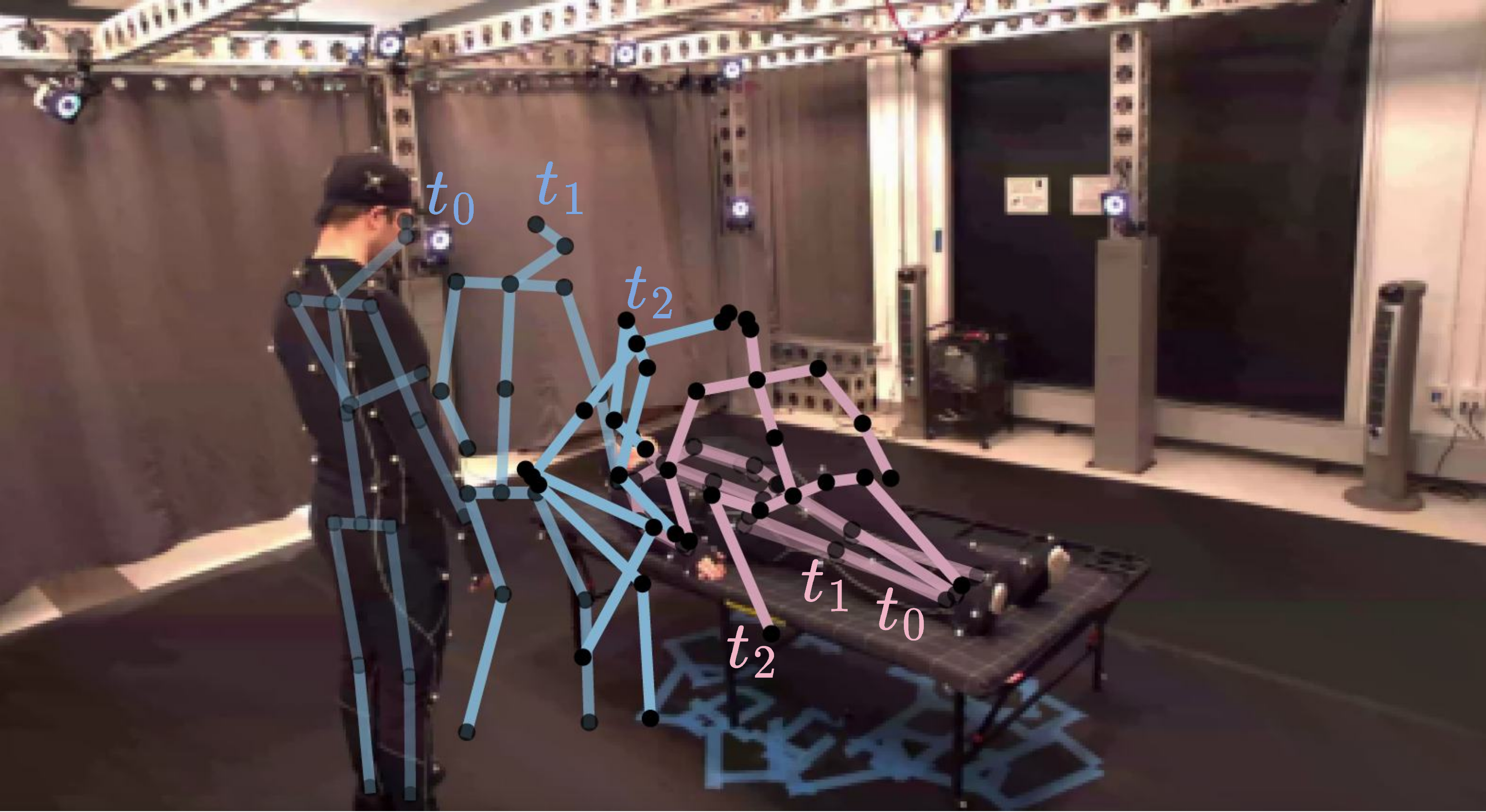}
    \caption{
    Predictions of joint positions for the caregiver (blue skeleton) and care receiver (pink skeleton) in a physical assistive task (lay-to-sit transfer), overlaid on a snapshot from a video clip. Future poses are illustrated for three different timesteps. $t_0$ represents the starting pose, while $t_1$ and $t_2$ correspond to two future timesteps with $0.5$ seconds of difference.}
  \label{fig:pull_figure2}
\end{figure}

\IEEEPARstart{T}{he} rising labor shortage and the increasing aging population drive the need for robots capable of assisting humans in need of care~\cite{padhan2023artificial, worth2024robots, aemoglu2021demographics}. A significant challenge in assisting humans, in contrast to object manipulation, is managing human agency. Ensuring safe physical interactions with humans requires robots to not only manipulate the care receiver's body but also anticipate their movements~\cite{kang2018real, mugisha2024motion}. 
Controllers or policies that lack awareness of human motion intention struggle to adapt swiftly to sudden changes in human actions~\cite{widmann2018human, mugisha2024motion}. Hence, the ability to predict human motion during physical interactions is vital for developing effective physically assistive robots. A challenge in motion prediction in these scenarios is the coupled dynamics that emerge from the physical interaction between two agents due to the reciprocal influence of one agent's actions on the other. Take, for example, the task of transferring an elderly person from a laying position to a sitting position (\Cref{fig:pull_figure2}). In this setting, the caregiver (CG) approaches the care receiver (CR) and initiates contact, which in turn signals the CR to engage their muscles for the lift. Through contact interaction, the CG senses the amount of effort exerted by the CR and applies sufficient force to support the maneuver. Here, the actions of the CR and CG are interleaved and influence each other. 

In this work, we study the problem of motion prediction in physically assistive tasks. Given the scarcity of physical human-robot interaction (HRI) data, we focus on human-human interaction scenarios, where collecting a large dataset is practical using well-established techniques like motion capture. We believe this data and the techniques presented in this work are transferable to physical HRI settings and reserve this for future study. 

We collected 908 clips from assistive tasks: sit-to-stand transfer from a chair, lay-to-sit transfer from a bed, lay-to-stand transfer from a bed, and unconstrained movements. Each task includes different sequences of actions or maneuvers. This dataset promises to be a valuable resource for (i) predicting human motion, (ii) data-driven methods that distill control policies for robots, (iii) performance baselines for physical human-robot interaction, and (iv) informing the design of robots that operate in assistive settings.

With this dataset in hand, we aim to predict human movements by predicting future pose sequences based on previously observed poses.
This area has attracted considerable interest due to its importance in applications such as autonomous driving~\cite{du2019ral}, human-robot collaboration~\cite{duarte2018action, vianello2021human}, and robotic navigation~\cite{chen2019crowd, chen2020relational}. 
The problem is inherently complex, requiring both spatial and temporal reasoning, and is compounded by the variability of assistance scenarios and human dynamics. In addition to the aforementioned challenges, physical interactions introduce emerging coupled dynamics, which add yet another layer of complexity.

Inspired by the success of Denoising Diffusion Probabilistic Models (DDPMs)~\cite{ho2020ddpm} in image generation~\cite{rombach2022high} and recently in human pose prediction~\cite{saadatnejad2023generic}, we extend these methods to motion interaction scenarios in physically assistive tasks. To this end, we propose an interaction-aware denoising diffusion (IDD) model capable of producing realistic and accurate predictions of human poses. To the best of our knowledge, this is the first pose prediction model that considers close contact interactions between agents in physical assistive tasks.

Our model predicts an agent's pose by conditioning not only on its own previous poses but also on the pose of the interacting agent, allowing for dynamic adjustments that reflect the nature of the interaction. An example use case of our model is illustrated in \Cref{fig:pull_figure2}, where the predicted joint positions of both the caregiver and the care receiver are shown for three representative timesteps. We validate the effectiveness of our approach through extensive experiments on our dataset.

In summary, our contributions are three-fold:
\begin{itemize}
    \item We present the HHI-Assist dataset, a collection of motion capture data capturing human-human interaction (HHI) for physical assistance.
    \item We propose an interaction-aware denoising diffusion (IDD) model that generates realistic and accurate predictions of human poses.
    \item We perform experiments to evaluate the performance of our model, assess its generalization, and investigate alternative representations for predicting human poses in interactive scenarios.
\end{itemize}

\section{Related Work}

While large-scale datasets for single-person motion, such as AMASS~\cite{amass2019} and Human3.6M~\cite{h36m}, are widely available, datasets capturing human-human interactions remain relatively limited.
These datasets can be categorized into the Social Interaction and Close Contact Interaction datasets.
Social Interaction datasets primarily focus on non-physical interactions, such as conversations and crowd navigation. For instance, the World-Pose Football dataset~\cite{jiang2024worldpose} and JRDB~\cite{martin2021jrdb} are utilized mainly for analyzing spatial dynamics and non-contact activities in sports and urban settings. 
Similarly, 3DPW~\cite{3dpw} and EXPI~\cite{expi} provide insights into multi-person interactions that do not involve direct physical contact, supporting research on weak social interactions~\cite{tanke2023social, adeli2020socially}.
On the other hand, Close Contact datasets~\cite{yin2023hi4d, fieraru2023reconstructing, ko2021air} capture fine-grained close-contact interactions that involve physical touch or close proximity.
Among these, Hi4D~\cite{yin2023hi4d} and AIR-Act2Act~\cite{ko2021air} employ markerless motion capture systems and CHI3D~\cite{fieraru2023reconstructing} uses a semi-marker-based setup in which participants alternate wearing marker suits.
Our dataset, HHI-Assist, expands on this category by providing the first marker-based motion capture data specifically designed for physical assistance scenarios. In contrast to existing datasets, HHI-Assist focuses on direct and strong physical interactions, offering a richer resource for studying assistive behaviors.

Predicting future positions of humans at a coarse-grained level, such as predicting center positions~\cite{bahari2025certified, saadatnejad2022sattack} or bounding boxes~\cite{saadatnejad2022pedestrian}, has been extensively studied. Our work, however, focuses on predicting the fine-grained human pose (parameterized as joint positions). Unlike other studies that incorporate additional context, such as action class~\cite{cai2021unified} or extra modalities~\cite{saadatnejad2024socialtransmotion}, we limit our focus to the observation pose sequence alone. 
Our prediction horizon ranges from a few hundred milliseconds to one second, which aligns with typical settings in receding horizon control for high-degree-of-freedom robotic systems~\cite{koenemann2015whole}.

In human pose prediction, early efforts used feed-forward networks~\cite{li2018convolutional}, and later Recurrent Neural Networks (RNNs) to model the temporal aspects of the task~\cite{fragkiadaki2015recurrent, martinez2017human}. Later advancements integrated Graph Convolutional Networks (GCNs) to better capture the spatial dependencies of human poses~\cite{mao2019learning, liu2021motion}. 
Subsequently, a unified GCN captured spatio-temporal features~\cite{sofianos2021space}, and a two-stage GCN refined predictions~\cite{ma2022progressively}.
More recently, Transformers have demonstrated effectiveness in capturing spatial and temporal dependencies, achieving state-of-the-art performance in human pose prediction. Various Transformer architectures have been explored, including serial spatial and temporal attention blocks~\cite{saadatnejad2024toward}, parallel spatial and temporal blocks~\cite{aksan2021spatio}, and hybrid models combining both approaches~\cite{zhu2022motionbert}. Our model also leverages the Transformer architecture to capture the spatio-temporal relationship in human pose sequences.

Generative models have lately been utilized to learn the distribution of human motions better. To this end, Generative Adversarial Networks (GANs)~\cite{kundu2019bihmp} and Variational Auto-Encoders (VAEs)~\cite{walker2017pose, cai2021unified} have been widely used due to their capability to learn complex data distributions. 
To promote diversity, various strategies of sampling have been proposed~\cite{yuan2020dlow,kim2024mdn}.
In addition, uncertainty in human pose prediction has been explored, including approaches that model homoscedastic uncertainty~\cite{saadatnejad2024toward}, as well as heteroscedastic uncertainty and multimodality through spatial heatmap representations~\cite{hosseininejad2025motionmap}.

Building on the success of diffusion models in image generation~\cite{rombach2022high,zhang2023adding}, these models have recently been applied to time-series imputation (i.e., missing value replacement)~\cite{tashiro2021csdi} and human pose prediction~\cite{chen2023humanmac, barquero2023belfusion, saadatnejad2023generic}. By effectively modeling the data distribution, diffusion models can generate poses that are both realistic and accurate. 
Our work advances this research direction by introducing a denoising diffusion model specifically tailored for interaction-aware pose prediction. We train and evaluate this model on our dataset and demonstrate its effectiveness in capturing complex interactive dynamics.

\section{Dataset}
\label{sec:dataset}

In this section, we present the HHI-Assist dataset.

\subsection{Data Collection}
Data was collected using an Optitrack Motion Capture rig with 20 Optitrack Prime 17W infrared tracking cameras and Motive 3.0 with the skeleton model~\cite{optitrack_prime17w}. This skeleton model includes 21 joints and 20 links, representing an average human. Link dimensions were calibrated to match those of the participants. 
Support equipment, such as the chair (43.18 L x 44.45 W x 46.99–54.61 H (cm)) and the bed (190.5 L x 99.06 W x 45.72 H (cm)), were also instrumented with markers.
Participants wore motion capture suits with 50 reflective markers~\cite{SkeletonMarkerSetCore50}.
Data is saved in files with a BVH format\cite{meredith2001motion} at a frequency of 120 Hz. 
Data were excluded upon manual inspection when the markers were occluded enough to cause unnatural or infeasible skeleton link overlap behavior (e.g., caregiver's arm crossing through the care receiver's arm).

Participants are healthy individuals from our laboratory who were not involved in the project, aged 21 to 50 years, with an average height of 169 ± 17 cm. Prior experience as a care receiver or caregiver was not required or explicitly considered when recruiting the participants. All participants agreed and signed informed consent and received a non-monetary incentive equivalent to 10 USD/h.
Pairs of individuals were drawn from our pool of participants. Each pair was given a live demonstration of the maneuver, acted out by the facilitators, along with a verbal description of the major steps. Participants were then assigned roles (caregiver or care receiver) and instructed to demonstrate the maneuvers.
The maneuvers were selected based on demonstrations by an occupational therapist showing various possible ways to assist care receivers with each task. Variations account for differences in care receiver's range of motion, strength, and areas of sensitivity or pain.
While the dataset reflects plausible motions used in care scenarios, it does not guarantee that these motions are ergonomically correct or safe for clinical use.

We ensure participant privacy by anonymizing identities and distributing only motion capture data, with all video footage excluded.

\begin{table}[!t]
    \centering
    \caption{Details of tasks, maneuvers, unique participant pairs, total takes / demonstrations in the HHI-Assist dataset.}
    \begin{tabular}{lllcc}
        \toprule
        \textbf{N} & \textbf{Task} & \textbf{Maneuvers} & \textbf{Pairs} & \textbf{Takes} \\
        \midrule
        & \multirow{3}{2.5cm}{Sit-to-stand transfer \\ from a chair} & Side Assist & 11 & 173 \\
        1&& Lever & 11 & 160 \\
        && Front Assist (Hug) & 11 & 167 \\
        \midrule
        &\multirow{3}{2.5cm}{Lay-to-sit transfer \\ from a bed} & Full Assist & 12 & 131 \\
        2&& Side Assist & 12 & 129 \\
        && Front Assist (Hug) & 13 & 129 \\
        \midrule 
        3&Lay-to-stand transfer & Full Assist & 1 & 10 \\
        \midrule 
        4&Unconstrained & N/A & 1 & 9 \\
        \bottomrule
    \end{tabular}

    \label{tab:dataset}
\end{table}

\subsection{Dataset Details}
\Cref{tab:dataset} summarizes our dataset, which comprises 908 demonstrations spanning four assistive tasks:

\begin{enumerate} 
    \item \textbf{Sit-to-stand transfer from a chair:} This task involves transferring the care receiver from a seated position on a chair to a stable standing posture with assistance from the caregiver.
    \item \textbf{Lay-to-sit transfer from a bed:} This task involves transferring the care receiver from a lying position on a bed to a stable sitting posture, assisted by the caregiver.
    \item \textbf{Lay-to-stand transfer from a floor:} This task involves the transfer of the care receiver from a supine position on the floor to a standing position with the caregiver's assistance, supported by a bed or chair. 
    \item \textbf{Unconstrained:} In this task, participants were instructed to pantomime or act out various scenarios with or without props. The scenarios include playing with a tennis ball, cooking in the kitchen, performing mirroring exercises (i.e., participants face each other and copy each other's movements without speaking), grooming and dressing tasks, dancing, item handover, ambulation assistance (from a seated position), fighting, and calisthenics/exercise (standing). 
\end{enumerate}

\begin{figure}[!t]
    \centering
    \includegraphics[width=\linewidth]{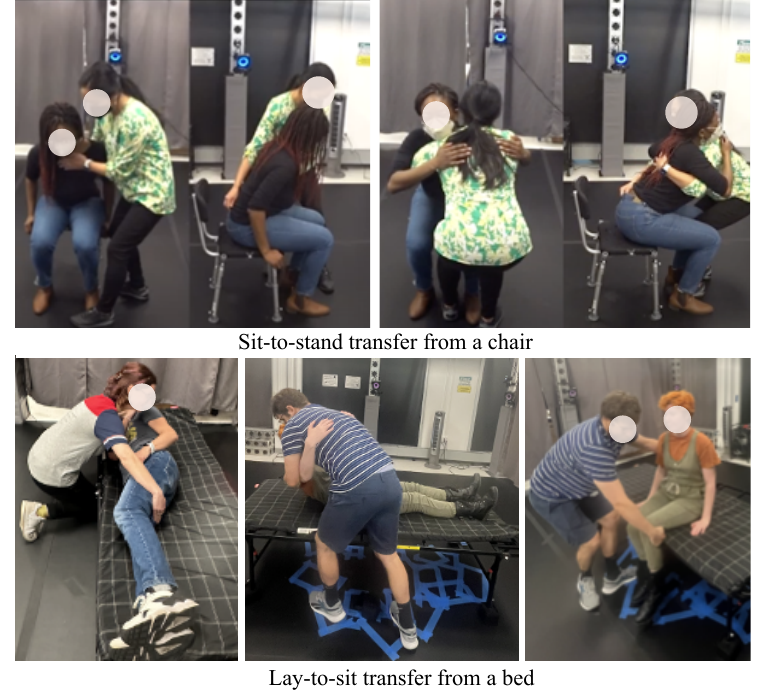}
    \caption{Examples of physical assistance scenarios from the HHI-Assist dataset. Top: Task 1 (Sit-to-Stand), with Side Assist on the left and Front Assist (Hug) on the right. Bottom: Task 2 (Lay-to-Sit), with Side Assist on the left and Full Assist in the center and on the right.}
  \label{fig:tasks}
\end{figure}

Tasks 1 and 2 constitute the core of the dataset, with 500 and 389 demonstrations respectively, each performed by 11 to 13 unique participant pairs across three maneuver variations. \Cref{fig:tasks} illustrates sample frames from them, showcasing different maneuver types. On average, sit-to-stand maneuvers lasted 9.4 seconds, while lay-to-sit transfers averaged 18.3 seconds.
Task 3 contains 10 demonstrations and is used exclusively for evaluating model generalization. Task 4 includes 9 demonstrations and serves as a source for initialization or data augmentation due to its fundamentally different motions.

\begin{figure}[!t]
    \centering
    \includegraphics[width=\linewidth]{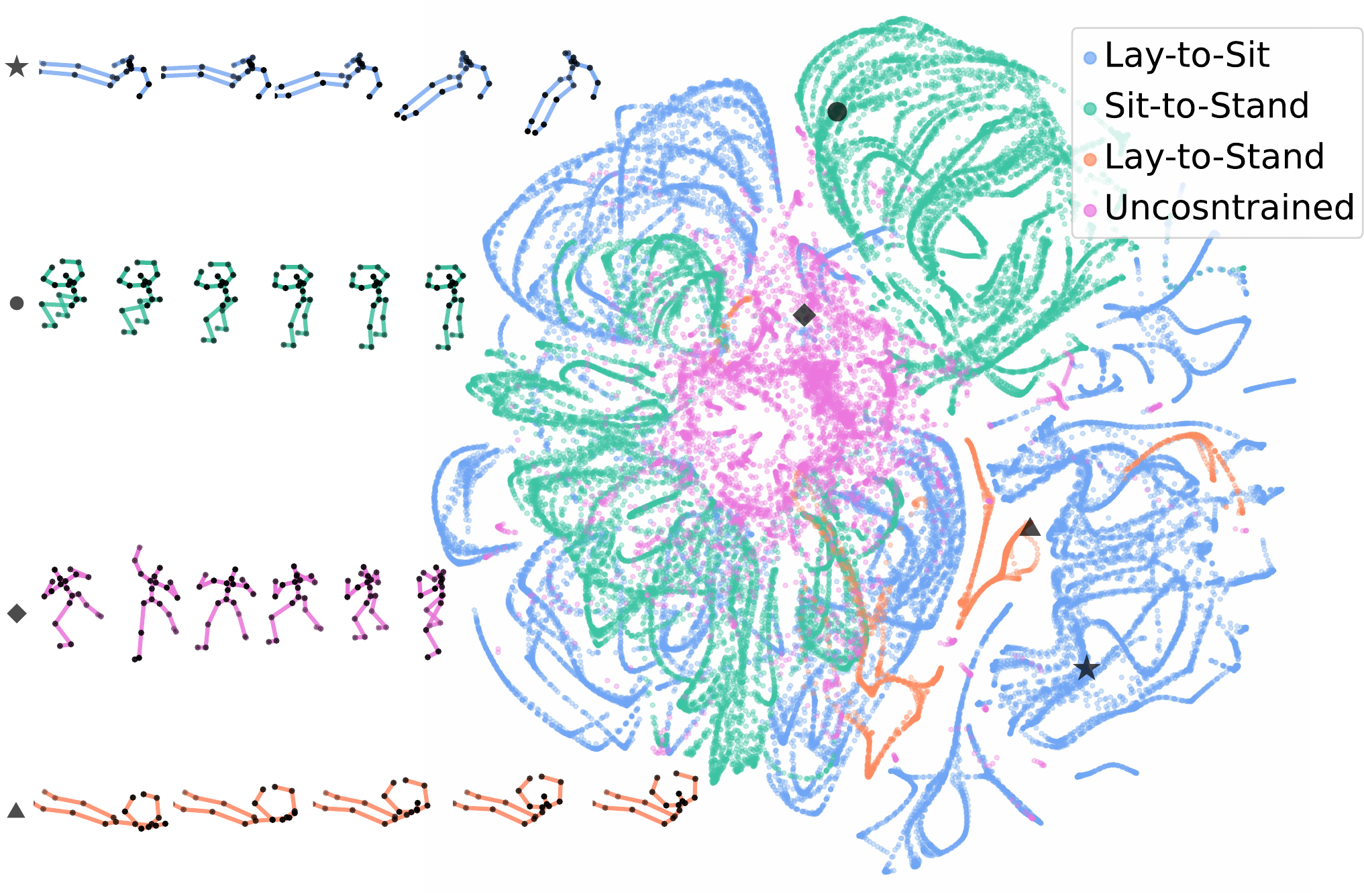}
\caption{The t-SNE plot of the HHI-Assist dataset motions showing the separability of the tasks, along with one randomly selected sample from each task.}
  \label{fig:tsne}
\end{figure}

The t-SNE visualization~\cite{vandermaaten2008tsne} of data motion sequences, along with four randomly selected samples, is presented in \Cref{fig:tsne}. The plot demonstrates that Task 1 and Task 2 are well-separated in the feature space, while Task 3 shows some overlap with both Task 1 and Task 2. This overlap is influenced by the clipped portion of data, which shares similarities with movements from either Task 1 or Task 2. Additionally, the unconstrained motions (Task 4) are relatively well-separated, which can be attributed to the distinct nature of these motions compared to the constrained tasks.

\section{Method}
\label{sec:method}
In this section, we begin by outlining the problem formulation and notations. Next, we detail the architecture of our model. Finally, we discuss the diffusion process in the training and inference of our Interaction-aware Denoising Diffusion (IDD) model.

\begin{figure*}[!t]
    \centering
    \includegraphics[width=\linewidth]{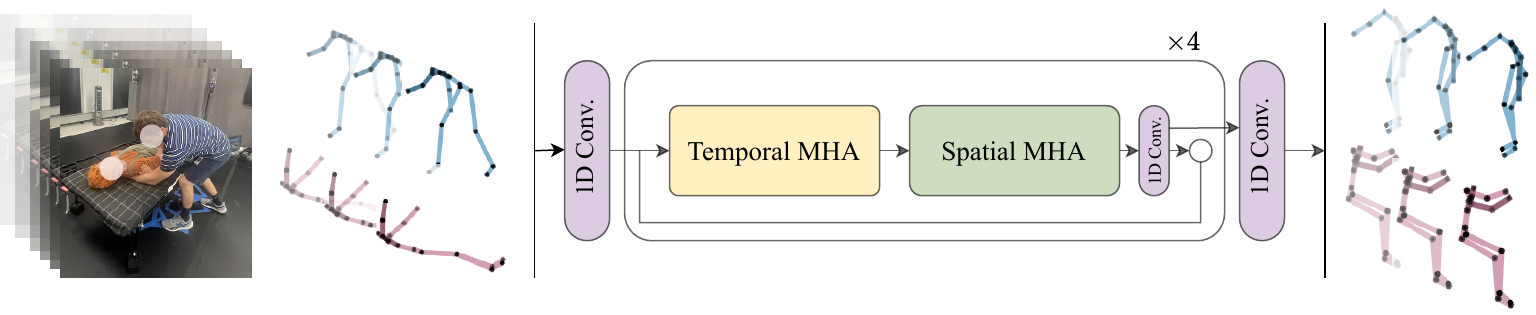}
    \caption{IDD takes the pose sequences of both CG (blue skeleton) and CR (pink skeleton) as input, predicting their corresponding future pose sequences, shown here for three timesteps. The architecture consists of two 1D convolution layers—one serving as an embedding layer for the inputs and another as the decoder—and four Transformer Blocks with skip connections. Each Transformer Block applies cascaded temporal and spatial multi-head attention to effectively capture spatio-temporal dependencies, followed by a 1D convolution layer. }
    \label{fig:model}
\end{figure*}

\subsection{Problem Formulation and Notations}

We denote the observed pose sequence of a subject $s \in \{\text{CG}, \text{CR}\}$, where $CG$ stands for the caregiver and $GR$ for care receiver, as 
\[
X_{s} = [p_{s_{-O+1}}, p_{s_{-O+2}}, \dots, p_{s_0}] \in \mathbb{R}^{O \times J \times 3},
\]
where $J$ is the number of joints, $O$ is the number of observed timesteps, and each joint pose $p$ is represented as a 3D point in Cartesian space.
The corresponding future pose sequence is
\[
Y_{s} = [p_{s_1}, \dots, p_{s_F}] \in \mathbb{R}^{F \times J \times 3},
\]
where $F$ is the number of timesteps to be predicted. 

In a high-level view, the model conditions on the concatenated observed sequences $[X_{\text{CG}}, X_{\text{CR}}]$ and predicts future poses $[Y_{\text{CG}}, Y_{\text{CR}}]$.

\subsection{Model Architecture}
We extend the Transformer model from previous work~\cite{saadatnejad2023generic} to address pose prediction in interactive scenarios. The model, as shown in \Cref{fig:model}, takes joint positions from two interacting agents and processes them through a 1D convolution layer and a series of residual transformer blocks. Each block applies temporal and spatial multi-head attention to capture the spatio-temporal dynamics of the sequences both between and within agents. The resulting embeddings from these blocks are summed and decoded with a 1D convolution layer to predict future poses. To effectively handle multivariate time series, each residual layer uses a two-dimensional attention mechanism: a temporal transformer layer learns dependencies across time, while a feature transformer layer models relationships among features.

\subsection{The Diffusion Process}
Each diffusion process has a noising and denoising part.
In the noising part, random Gaussian noise is gradually added to the clean data, transforming it into a pure Gaussian distribution with zero mean and identity covariance after $T$ steps. Mathematically, this can be expressed as: 
\[
x^t = \sqrt{\alpha_t} x^0 + \sqrt{1 - \alpha_t} \epsilon,
\]
where \( x^{t} \) is the noisy data at step \( t \), \( x^{0} \) is the original clean data, \(\alpha_t\) is a time-dependent scaling factor, and \(\epsilon \sim \mathcal{N}(0, I)\) is the Gaussian noise. 
In the denoising part, the model learns to denoise \( x^{T} \) and retrieve the original clean data \( x^{0} \).

In this work, we focus on the conditional diffusion process where a model can take the extra condition for a more accurate prediction. We tailor this process for our interaction-aware model that receives the observations of both agents as the conditions and is aimed to learn two conditional distributions 
\[
p(\Tilde{Y}_{CG}^{t}|X_{CG},X_{CR}) \quad \text{and} \quad p(\Tilde{Y}_{CR}^{t}|X_{CG},X_{CR}).
\]
During training, the model takes a sample of the embedded noisy pose sequences at a diffusion step $t$ by 
\[ 
\Tilde{Y}_{s}^t = \sqrt{\alpha_t} Y_{s} + \sqrt{1 - \alpha_t} \epsilon,
\]
 as input and learns to predict the added noise at that step $\epsilon_\theta$.
The loss function can be written as:
\begin{align*}
  \mathcal{L} = 
  \mathbb{E}_{Y_{s}, \epsilon, t}
  \| \epsilon - \epsilon_\theta (\Tilde{Y}_{s}^{t}, t \mid X_{CG}, X_{CR}) \|_2^2 ,
  \label{eq:loss_cond_adapted}
\end{align*}
where $Y_{s}$ refers to clean data and $\theta$ indicates learnable parameters of the model.

During inference, we begin with a sample drawn from a Gaussian distribution and iteratively denoise it using the trained network, conditioned on the interacting observation. This step-by-step process produces a noise sequence for CG and CR. 
In the final step of the diffusion process, the model generates the pose sequences by subtracting the last estimated noise from the last noisy input resulting in the predicted pose sequence $\Tilde{Y}_{CG}^{0}$ and $\Tilde{Y}_{CR}^{0}$. 
In summary, the network's noise estimations define a trajectory that progressively transforms the initial noise into the pose distribution, guided by the interacting observations.  
In our experiments, we set the diffusion steps $T=50$.

\section{Experiments}

\subsection{Dataset and Metric}

We use the first two tasks of the HHI-Assist dataset (sit-to-stand transfer from a chair and lay-to-sit transfer from a bed) for training and testing and consider the third task (lay-to-stand) only for evaluating the generalization of the models. We use data from the fourth task (unconstrained) for the initialization of the models.
After downsampling the video clips to a frame rate of 24 fps, each sequence in the dataset contains $O=24$ observation timesteps and $F=24$ prediction timesteps, corresponding to 1 second of observed and 1 second of future motion data. 
The dataset is divided into a training set with $44.8k$ sample sequences, a validation set with $3.5k$ sequences, and a test set with $8.7k$ sequences, with no participant overlap between the test set and the others.

We evaluate accuracy using the Mean Per Joint Position Error (MPJPE) measured in millimeters (mm) per timestep, as well as the overall average MPJPE across all timesteps. This metric averages the Euclidean distance between each predicted keypoint and the ground truth pose for all joints and is calculated after aligning the base joint (pelvis).

\subsection{Baselines}

In addition to our Interaction-aware Denoising Diffusion (IDD) model, we implemented the following learning-based and non-learning-based baselines:

\begin{enumerate}
    \item SiMLPe~\cite{guo2022back}: A competitive multi-layer perceptron (MLP) network designed for pose prediction.
    \item TCD~\cite{saadatnejad2023generic}: A denoising diffusion model for single human pose prediction.
    \item DSTFormer~\cite{zhu2022motionbert}: A dual-stream spatio-temporal Transformer encoder for learning efficient representations of human poses, adapted here for pose prediction.
    \item \textit{Zero-Vel}: A baseline that predicts future poses by assuming no movement, outputting the last observed pose for all future timesteps.
    \item \textit{Constant-Vel}: A baseline that assumes constant velocity, predicting future poses by calculating it from the last two observed poses and extrapolating forward.
\end{enumerate}

\subsection{Implementation details}
Our model was trained using the Adam optimizer with an initial learning rate of 1e-3, which was reduced by a factor of 0.1 after 75\% and 90\% of the 50 total epochs. Training was conducted on a single NVIDIA GeForce RTX 3090 GPU (24GB VRAM) and took approximately one day to complete.

\subsection{Results}

We present the quantitative results of our model in comparison to the baselines for predicting the poses of both the caregiver and the care receiver at various prediction horizons in \Cref{tab:res}.
We first observe that predicting the pose of the care receiver is generally easier for all models, as the care receiver exhibits less movement compared to the caregiver. 
Second, we observe that the interaction-aware model (IDD) significantly outperforms the interaction-unaware models and other baselines for both caregiver and care receiver, highlighting the importance of capturing interaction dynamics and demonstrating its effectiveness.

\begin{figure*}[!t]
    \centering
    \includegraphics[width=0.95\textwidth]{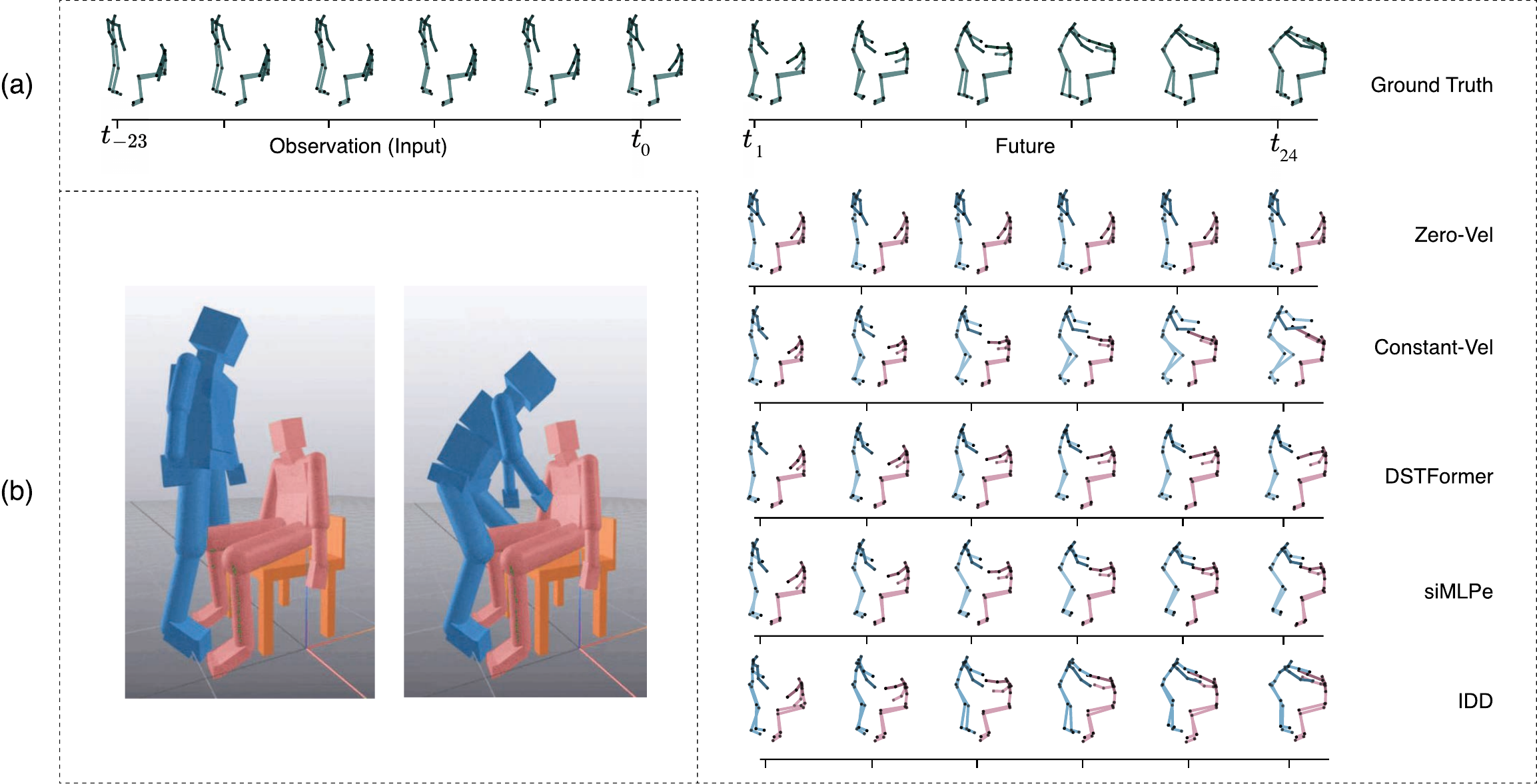}
    \caption{(a) Qualitative comparison of pose predictions from the baselines and our model on the HHI-Assist dataset shown every four timesteps for both caregiver (blue skeleton) and care receiver (pink skeleton) put together. The models take as input observation the pose skeleton from $t_{-23}$ to $t_0$ (depicted in the top row in green skeleton) and are supposed to predict future poses close to the ground truth for $t_1$ to $t_{24}$. We can observe that IDD predicts poses more accurately. (b) Visualization of two example frames of assisting scenarios on the Drake simulator.}
    \label{fig:bl}
\end{figure*}

\Cref{fig:bl} (a) shows a qualitative comparison of predictions from IDD and the baselines, along with the observation input and ground truth future poses. The models take as input the pose skeleton from $t_{-23}$ to $t_0$ and are expected to predict future poses close to the ground truth from $t_1$ to $t_{24}$. The results reveal that IDD effectively captures the data distribution, producing pose predictions that are realistic, close to the ground truth, and outperform the baselines. 
We have also provided a 3D visualization of two example frames in \Cref{fig:bl} (b), generated using the Drake simulator~\cite{drake}.

\begin{table*}[!ht]
    \centering
    \caption{Quantitative comparison of pose predictions from the baselines and our model on the HHI-Assist dataset for both caregiver and care receiver in terms of MPJPE ($\mathrm{mm}$) at different prediction horizons. Best performing values across each column are shown in bold.}
    \begin{tabular}{lcccccc|cccccc}
        \toprule
        & \multicolumn{6}{c}{Caregiver} & \multicolumn{6}{c}{Care receiver} \\
        Model & $85\,\mathrm{ms}$ & $330\,\mathrm{ms}$ & $580\,\mathrm{ms}$ & $750\,\mathrm{ms}$ & $1000\,\mathrm{ms}$ & average & $85\,\mathrm{ms}$ & $330\,\mathrm{ms}$ & $580\,\mathrm{ms}$ & $750\,\mathrm{ms}$ & $1000\,\mathrm{ms}$ & average \\
        \midrule
        Constant-Vel & \textbf{5.2}  & 40.5  & 89.3  & 124.1 & 176.6 & 80.6 &
        \textbf{5.4}  & 33.7  & 72.8 & 100.9 & 143.9 & 66.0 \\
        Zero-Vel & 14.5 & 64.0  & 92.9  & 109.3 & 123.8& 72.7& 10.8 & 39.0  & 62.8  & 76.7  & 95.4 & 54.7 \\

        DSTFormer~\cite{zhu2022motionbert} & 8.5 & 32.7 & 62.7& 81.2 & 105.0 & 54.8 
        & 8.3 & 24.4 & 44.1& 57.3 & 77.9 & 39.9 \\
        siMLPe~\cite{guo2022back} & 6.9 & 30.9 & 58.9 & 76.8 & 100.2 & 51.7 &
        6.5 & 23.1 & 43.0 & 56.0 & 75.7 & 38.5\\

        TCD~\cite{saadatnejad2023generic} & 6.9  & 31.8  & 60.2  & 78.1 & 100.1 & 52.0 & 6.4  & 23.9  & 42.8 & 55.5 & 73.9 & 37.5 \\
        \midrule
        IDD    & 6.1  & \textbf{29.8}  & \textbf{58.6}  & \textbf{74.6}  & \textbf{94.0} & \textbf{50.4}& 5.5  & \textbf{21.7}  & \textbf{39.6}  & \textbf{49.5}  & \textbf{63.2}  & \textbf{34.3} \\
        \bottomrule
    \end{tabular}
    \label{tab:res}
\end{table*}

\subsection{Discussions}

\subsubsection{Generalization}
Here, we assess the generalizability of our model to unseen state distribution not included in the training dataset. Specifically, we examine whether our model trained on Task 1 and Task 2, can effectively generalize to Task 3. When evaluated on Task 3, IDD achieves an average MPJPE of $89.3$ mm and a long-term MPJPE error at 1000 ms of $154.8$ mm for CG and an average MPJPE of $62.5$ mm and a long-term MPJPE error at 1000 ms of $113.8$ mm for CR. 
It demonstrates the model's ability to adapt to novel tasks, though improving performance under such distribution shifts remains a direction for future work.

\subsubsection{Delayed Coupled Dynamics}
Modeling the coupled dynamics between agents can be challenging in certain scenarios. 
Here, we investigate the limit of information transfer between agents, i.e., to determine how much predictive power can be gained if we have prior knowledge of the other person’s motion.
In this experiment, without collecting new data, we simulate a scenario where one agent waits for $0.5$ seconds before responding to the other's movement.
Our modified model, called \textquotedblleft Delayed IDD\textquotedblright, takes as input the observation sequence of CG/CR from $t=0$ to $t=1$ and a delayed observation of CR/CG from $t=0.5$ to $t=1.5$, and is supposed to predict $t=1$ to $t=2$ of CG/CR. This allows each agent to provide a better movement, simplifying the prediction task by reducing the immediate complexity of mutual interaction.
Delayed IDD achieves an average MPJPE of $47.6$ mm and a long-term MPJPE at 1000 ms of $89.9$ mm for CG, and $32.5$ mm and $60.0$ mm, respectively, for CR.
The enhanced prediction accuracy for both agents indicate that the Delayed IDD model can better anticipate each agent's subsequent movements by decoupling their interactions to some extent.

\subsubsection{Other Representations}

In addition to joint positions, we study the impact of using joint angles as an alternative representation. Joint angles describe the relative orientation between connected body links, offering a potentially more compact and invariant representation of human motion. They can be particularly advantageous when maintaining fixed link lengths is important.
However, discontinuities in some angular representations, such as Euler angles and quaternions, make them difficult to learn~\cite{saxena2009learning, zhou2019continuity}. In this work, we focus on rotation matrices, though they are not the most compact representation. We leave exploration of alternative angle-based representations for future work.

\begin{figure}[!t]
    \centering
    \includegraphics[width=0.95\linewidth]{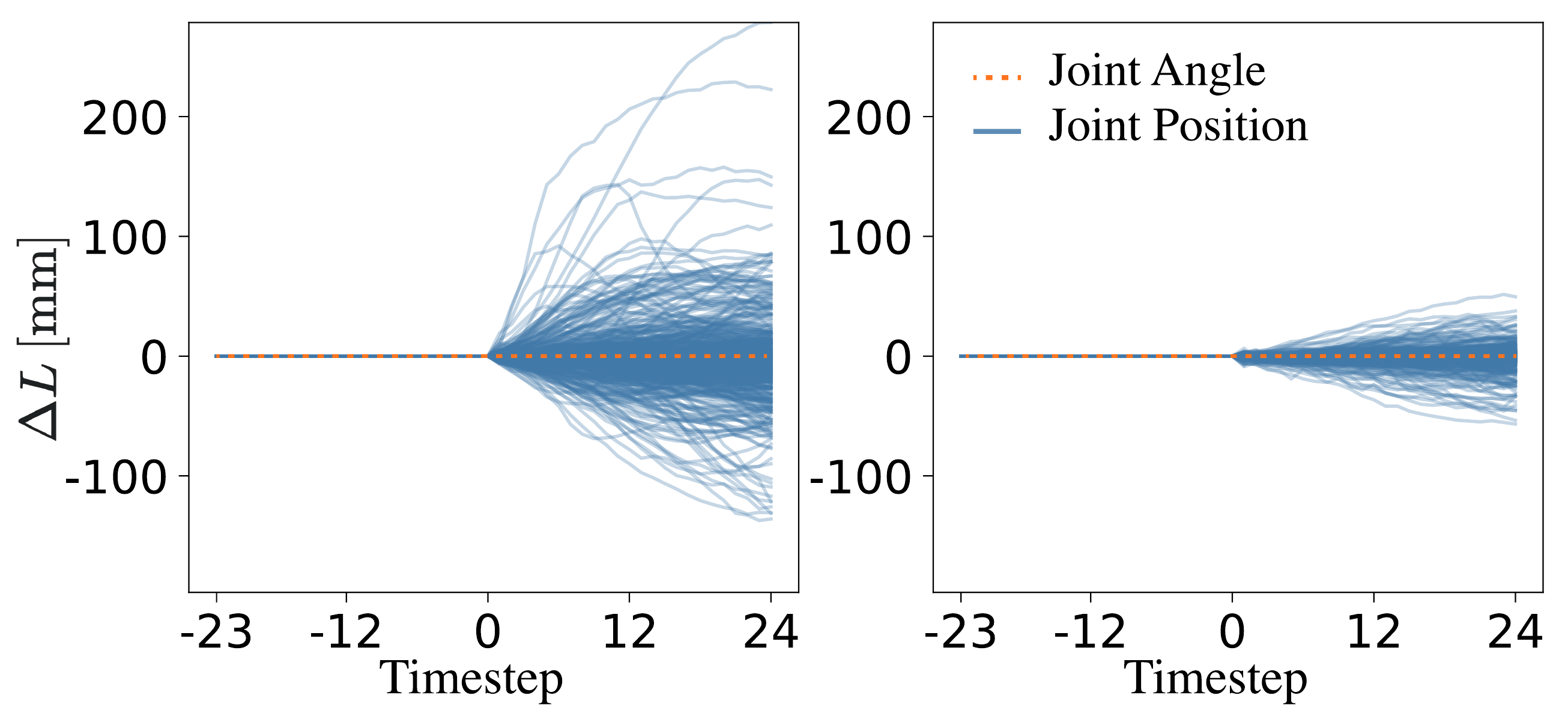}
    \caption{Comparison of link length variations in predictions using joint position and joint angle representations. The plots show changes in link lengths across different timesteps for $500$ randomly selected motions. The left plot refers to the forearm, and the right plot refers to the neck. Predicting joint positions results in link length variations, whereas predicting joint angles preserves consistent link lengths.
    }
  \label{fig:links_lengths}
\end{figure}

To compare representations, we trained our model twice—once using joint positions and once using joint angles—and evaluated their prediction accuracies. 
As shown in \Cref{tab:rep_effect}, the model trained with joint positions achieved better performance in terms of MPJPE, , likely due to better alignment with the loss and evaluation metric. In contrast, the model trained with joint angle representation showed around $7\%$ higher average MPJPE but ensured consistent link lengths throughout the predictions.
Note that by applying the forward kinematics, we verified that the predicted rotation matrices are members of the SO(3) manifold, i.e., valid rotation matrices. 
For the joint position-based model, the mean absolute and standard deviation of link length changes across the test set were 7.1 mm and 14.84 mm, respectively. \Cref{fig:links_lengths} shows length variation of two representative links across all timesteps (both observed and predicted) for $500$ random examples. These results confirm that joint positions lead to variations in link lengths, with some samples exhibiting significant changes, whereas joint angles preserve consistent link lengths.

\begin{table}[!t]
    \centering
    \caption{Comparing the performance of IDD given joint position and joint angle representations in terms of MPJPE ($\mathrm{mm}$) at $1000\,\mathrm{ms}$ and average across all prediction horizons.}
    \begin{tabular}{l|cc|cc}
        \toprule
        & \multicolumn{2}{c}{Caregiver} & \multicolumn{2}{c}{Care receiver} \\
        Representation & $1000\,\mathrm{ms}$ & average & $1000\,\mathrm{ms}$ & average \\
        \midrule
        Joint Position & 94.0 & 50.4 & 63.2 & 34.3\\
        Joint Angle &  99.9 & 54.0 & 67.6 & 36.8 \\
        \bottomrule
    \end{tabular}
    \label{tab:rep_effect}
\end{table}

\section{Conclusions and Future Works} 
\label{sec:conclusion}

In this work, we presented the HHI-Assist dataset, a collection of caregiver and care receiver demonstrations in physical assistance scenarios. We proposed an interaction-aware denoising diffusion model for predicting human poses, achieving significant improvements in prediction accuracy by accounting for the dynamic interactions between caregivers and care receivers. We hope our work paves the way for more advanced interaction-aware motion prediction, ultimately enhancing downstream tasks such as robotic assistance policies and improving care for individuals in need of support.

Future work will focus on exploring alternative input representations and addressing various sources of uncertainty in the task. We also plan to investigate the model's potential for robot control, such as integrating predictions into receding horizon controllers (e.g., model predictive controllers) or incorporating them into the observation space for learned policies.
Beyond pose prediction, the HHI-Assist dataset enables new research avenues, particularly in policy learning for physical human-robot interaction. The dataset can be retargeted to robot kinematics, providing valuable training data for behavior cloning techniques such as Diffusion Policy~\cite{chi2023diffusionpolicy}. It can also serve as a source of style data for reinforcement learning methods, such as Adversarial Motion Priors~\cite{peng2021amp}. Lastly, the dataset allows for deriving performance metrics for assistive tasks, including evaluations of robot performance in terms of contact sequences, force ranges, and kinematic limits.

\section*{Acknowledgments}
\noindent The authors would like to thank Eric Dusel, Bisi Chikwendu, and  Andrew Silva of Toyota Research Institute, and Saged Bounekhel of EPFL. E.D. and B.C. assisted with data collection, A.S. provided valuable feedback, and S.B. contributed to preliminary experiments.

\bibliographystyle{ieee_fullname}
\bibliography{references}

\end{document}